\documentclass[11pt,a4paper]{article}
\usepackage[hyperref]{emnlp-ijcnlp-2019}
\usepackage{times}
\usepackage{latexsym}
\usepackage{times}
\usepackage{fourier} 
\usepackage{array}
\usepackage{caption}
\usepackage{makecell}
\newcommand{\datasetname}{\textsc{PrivacyQA}}
\newenvironment{tight_enumerate}{
\begin{enumerate}
  \setlength{\itemsep}{0pt}
  \setlength{\parskip}{0pt}
}{\end{enumerate}}

\usepackage{enumitem}
\usepackage{multirow}
\usepackage{graphicx}
\usepackage{amssymb}
\usepackage{pifont}
\usepackage{url}
\usepackage{booktabs}
\newcommand{\cmark}{\ding{51}}%
\newcommand{\xmark}{\ding{55}}%
\usepackage{hyperref}
\usepackage{url}

\aclfinalcopy 

\newcommand\cmu{$^\diamondsuit$}
\newcommand\penn{$^\heartsuit$}
\newcommand\fordham{$^\spadesuit$}
\newcommand\aspace{\hspace{.75em}}

\title{Question Answering for Privacy Policies:\\ Combining Computational and Legal Perspectives}

\author{
 Abhilasha Ravichander \cmu \aspace 
 Alan Black \cmu \aspace
 Shomir Wilson \penn \aspace \\
 \textbf{Thomas Norton}\fordham \aspace
 \textbf{Norman Sadeh}\cmu \\
 \cmu Carnegie Mellon University, Pittsburgh, PA\\
 \penn Penn State University, University Park, PA
 \fordham Fordham Law School, New York, NY\\
  {\tt \{aravicha, awb, sadeh\}@cs.cmu.edu} \\
  {\tt \{shomir\}@psu.edu}, {\tt \{tnorton1\}@law.fordham.edu} 
}

\date{}

\begin{document}
\maketitle
\begin{abstract}
Privacy policies are long and complex documents that are difficult for users to read and understand, and yet, they have legal effects on how user data is collected, managed and used.  Ideally, we would like to empower users to inform themselves about issues that matter to them, and enable them to selective explore those issues. We present \datasetname, a corpus consisting of 1750 questions about the privacy policies of mobile applications, and over 3500 expert annotations of relevant answers. We observe that a strong neural baseline underperforms human performance by almost 0.3 F1 on \datasetname, suggesting considerable room for improvement for future systems. Further, we use this dataset to shed light on challenges to \emph{question answerability}, with domain-general implications for any question answering system.  The \textsc{PrivacyQA} corpus offers a challenging corpus for question answering, with genuine real-world utility.

\end{abstract}

\section{Introduction}
Privacy policies are the documents which disclose the ways in which a company gathers, uses, shares and manages a user's data. As legal documents, they function using the principle of \emph{notice and choice} \citep{federal1998privacy}, where companies post their policies, and theoretically, users read the policies and decide to use a company's products or services only if they find the conditions outlined in its privacy policy acceptable. Many legal jurisdictions around the world accept this framework, including the United States and the European Union \cite{patrick1980privacy, oecd2004oecd}. However, the legitimacy of this framework depends upon users actually reading and understanding privacy policies to determine whether company practices are acceptable to them \citep{reidenberg2015disagreeable}. In practice this is seldom the case \citep{cate2010limits, cranor2012necessary, schaub2015design,  gluck2016short, jain2016big, us2012protecting, mcdonald2008cost}. 
This is further complicated by the highly individual and nuanced compromises that  users are willing to make with their data \cite{leon2015privacy}, discouraging a `one-size-fits-all' approach to notice of data practices in privacy documents.

\begin{figure}[tb]
\centering
\includegraphics[scale=0.37]{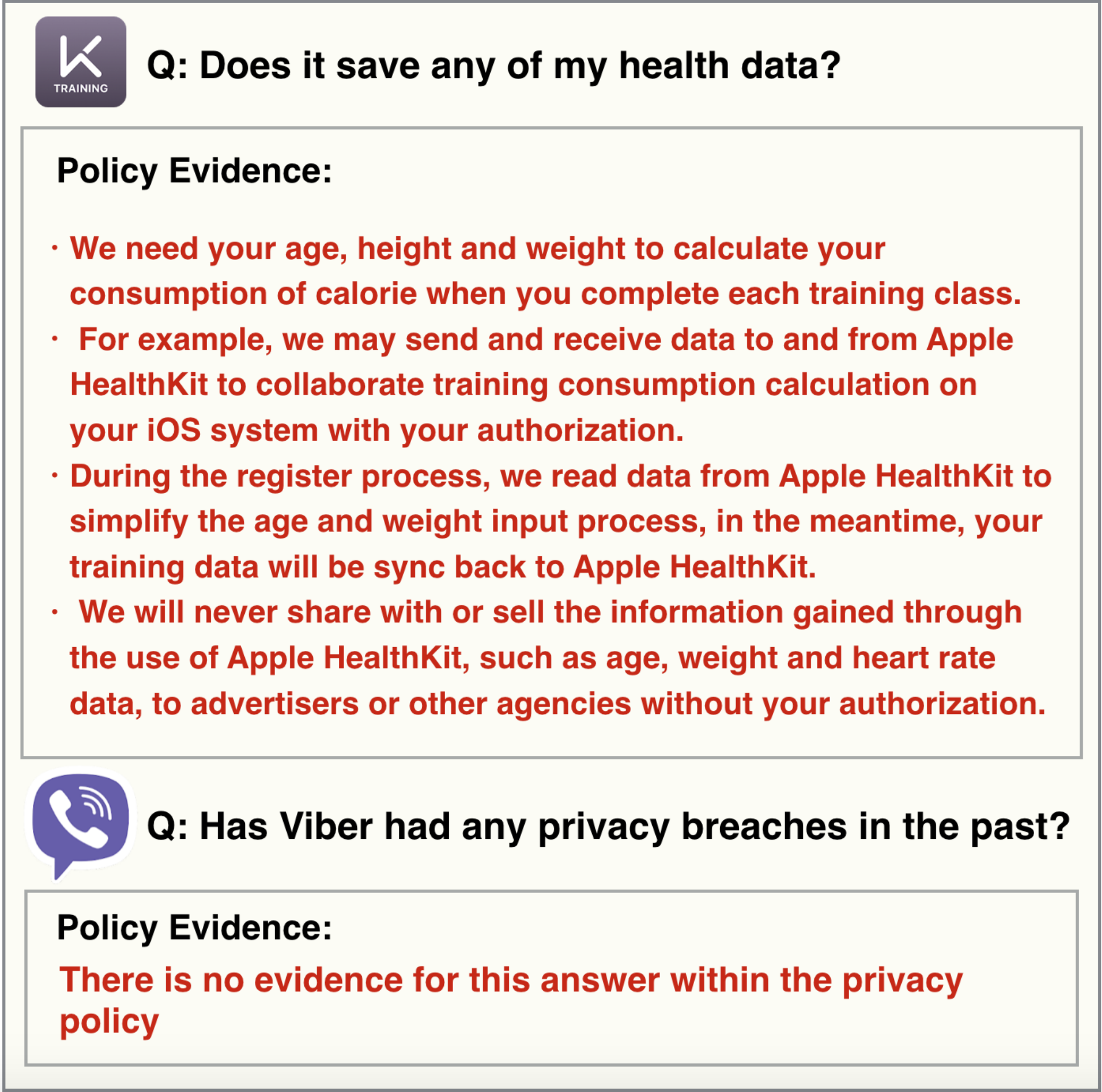}
\captionsetup{belowskip=0pt}
\caption{Examples of privacy-related questions users ask, drawn from two mobile applications: Keep\footnotemark  and Viber.\footnotemark  Policy evidence represents sentences in the privacy policy that are relevant for determining the answer to the user's question.\footnotemark}
\label{fig:results1}
\end{figure}

With devices constantly monitoring our environment, including our personal space and our bodies, lack of awareness of how our data is being used easily leads to problematic situations where users are outraged by information misuse, but companies insist that users have consented. The discovery of increasingly egregious uses of data by companies, such as the scandals involving Facebook and Cambridge Analytica \citep{cadwalladr2018revealed}, have further brought public attention to the privacy concerns of the internet and ubiquitous computing. This makes privacy a well-motivated application domain for NLP researchers, where advances in enabling users to quickly identify the privacy issues most salient to them can potentially have large real-world impact.



\footnotetext[1]{\url{https://play.google.com/store/apps/details?id=com.gotokeep.keep.intl}}
\footnotetext[2]{\url{https://play.google.com/store/apps/details?id=com.viber.voip}}
\footnotetext[3]{A question might not have any supporting evidence for an answer within the privacy policy.}

Motivated by this need,  we contribute \datasetname, a corpus consisting of 1750 questions about the contents of privacy policies\footnote{All privacy policies in this corpus are in English.}, paired with over 3500 expert annotations. The goal of this effort is to kickstart the development of question-answering methods for this domain, to address the (unrealistic) expectation that a large population should be reading many policies per day. In doing so, we identify several understudied challenges to our ability to answer these questions, with broad implications for  systems seeking to serve users' information-seeking intent. By releasing this resource, we hope to provide an impetus to develop systems capable of language understanding in this increasingly important domain.\footnote{\textsc{PrivacyQA} is freely available at \url{https://github.com/AbhilashaRavichander/PrivacyQA_EMNLP}.} 

\begin{table*}[t]
\small
\centering
\begin{tabular}
{c|c|c|c|c|c|c}
\toprule
\multirow{2}{*}{Dataset} &\multirow{2}{*}{\parbox{1cm}{\centering Document Source}} &\multirow{2}{*}{\parbox{1.5cm}{\centering Expert Annotator}} & 
\multirow{2}{*}{\parbox{1.5 cm}{\centering Simple Evaluation}}  & \multirow{2}{*}{\parbox{1.7cm}{\centering Unanswerable Questions}}  & \multirow{2}{*}{\parbox{2cm}{\centering Asker 
Cannot See Evidence}} \\
&&&&&\\\midrule
PrivacyQA & Privacy Policies  & \cmark & \cmark & \cmark & \cmark \\ \midrule
NarrativeQA~\cite{kovcisky2018narrativeqa} & Fiction  & \xmark & \xmark & \xmark & \cmark \\
InsuranceQA~\cite{feng2015applying} & Insurance  & \cmark & \cmark & \xmark & \cmark \\
TriviaQA~\cite{joshi2017triviaqa}& Wikipedia & \xmark & \cmark & \xmark & \cmark\\
SQuAD 1.0~\cite{rajpurkar2016squad} & Wikipedia  &  \xmark &  \cmark & \xmark & \xmark\\
SQuAD 2.0~\cite{rajpurkar2018know}& Wikipedia  &  \xmark &  \cmark & \cmark & \xmark\\
MS Marco~\cite{nguyen2016ms}& Web Documents  & \xmark & \xmark & \cmark & \cmark\\
MC Test~\cite{richardson2013mctest}& Fiction & \xmark & \cmark & \xmark & \xmark\\
NewsQA~\cite{trischler2016newsqa}& News Articles  & \xmark &  \cmark & \cmark & \cmark \\

\bottomrule 
\end{tabular}
\caption{Comparison of the \textsc{PrivacyQA} dataset to other question answering datasets. Expert annotator indicates domain expertise of the answer provider. Simple evaluation indicates the presence of an automatically calculable evaluation metric. Unanswerable questions indicates if the respective corpus includes unanswerable questions. `Asker Cannot See Evidence' indicates that the asker of the question was not shown evidence from the document at the time of formulating questions.} 
\vspace{-10pt}
 \label{tab:dataset_comparison}
\end{table*}

\section{Related Work}
Prior work has aimed to make privacy policies easier to understand. Prescriptive approaches towards communicating privacy information \cite{kelley2009nutrition, micheti2010fixing, cranor2003p3p} have not been widely adopted by industry. Recently, there have been significant research effort devoted to understanding privacy policies by leveraging NLP techniques \citep{sadeh2013usable, liu2016analyzing, oltramari2016privonto, mysoresathyendra-EtAl:2017:EMNLP2017, wilson2017analyzing}, especially by identifying specific data practices within a privacy policy. We adopt a personalized approach to understanding privacy policies, that allows users to query a document and selectively explore content salient to them. Most similar is the PolisisQA corpus \cite{harkous2018polisis}, which examines questions users ask corporations on Twitter. Our approach differs in several ways: 1) The \textsc{PrivacyQA} dataset is larger, containing 10x as many questions and answers. 2) Answers are formulated by domain experts with legal training. \footnote{This choice was made as privacy policies are legal documents, and require careful expert understanding in order to be interpreted correctly.} 3) \textsc{PrivacyQA} includes diverse question types, including unanswerable and subjective questions. 

Our work is also related to reading comprehension in the open domain, which is frequently based upon Wikipedia passages \citep{rajpurkar2016squad,rajpurkar2018know, joshi2017triviaqa, choi2018quac} and news articles \citep{trischler2016newsqa, hermann2015teaching, onishi-EtAl:2016:EMNLP2016}. Table.\ref{tab:dataset_comparison} presents the desirable attributes our dataset shares with past approaches. This work is also tied into research in applying NLP approaches to legal documents \cite{monroy2009nlp, quaresma2005question, do2017legal, kim2015applying, liu2015predicting, molla2007question, frank2007question}. While privacy policies have legal implications, their intended audience consists of the general public rather than individuals with legal expertise. This arrangement is problematic because the entities that write privacy policies often have different goals than the audience. \newcite{feng2015applying, tan-EtAl:2016:P16-1} examine question answering in the insurance domain, another specialized domain similar to privacy, where the intended audience is the general public.

\section{Data Collection}
\label{section:datacollection}






We describe the data collection methodology used to construct \textsc{PrivacyQA}. With the goal of achieving broad coverage across application types, we collect privacy policies from 35 mobile applications representing a number of different categories in the Google Play Store.\footnote{We choose categories that occupy atleast a 2\% share of all application categories on the Play Store \citep{story2018apps}}\footnote{As of April 1, 2018} One of our goals is to include both policies from well-known applications, which are likely to have carefully-constructed privacy policies, and lesser-known applications with smaller install bases, whose policies might be considerably less sophisticated. Thus, setting 5 million installs as a threshold, we ensure each category includes applications with installs on both sides of this threshold.\footnote{The final application categories represented in the corpus consist of books, business, education, entertainment, lifestyle, music, health, news, personalization, photography, productivity, tools, travel and game applications.} All policies included in the corpus are in English, and were collected before April 1, 2018, predating many companies' GDPR-focused \cite{voigt2017eu} updates. We leave it to future studies \cite{gallecase} to look at the impact of the GDPR (e.g., to what extent GDPR requirements contribute to making it possible to provide users with more informative answers, and to what extent their disclosures continue to omit issues that matter to users). 

\begin{figure}
\centering
\includegraphics[scale=0.49]{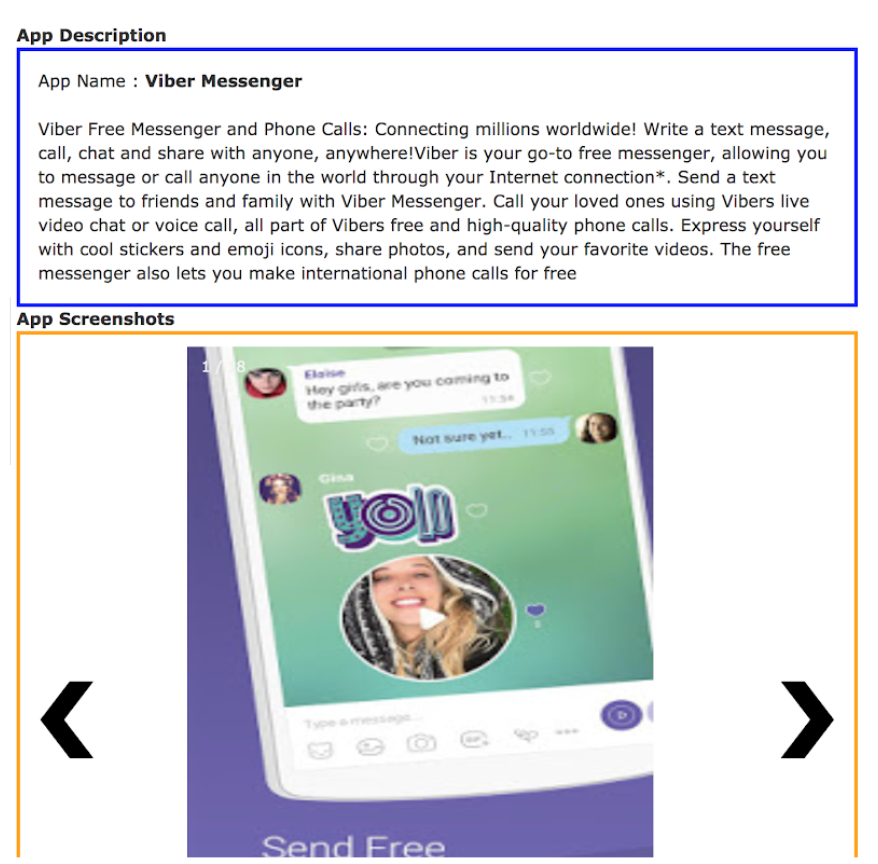}
\caption{User interface for question elicitation.}
\label{fig:screenshots-question}
\end{figure}
\subsection{Crowdsourced Question Elicitation}

The intended audience for privacy policies consists of the general public. This informs the decision to elicit questions from crowdworkers on the contents of privacy policies. We choose not to show the contents of privacy policies to crowdworkers, a procedure motivated by a desire to avoid inadvertent biases \cite{weissenborn-etal-2017-making, kaushik-lipton-2018-much, poliak-etal-2018-hypothesis, gururangan-etal-2018-annotation, naik-etal-2018-stress}, and encourage crowdworkers to ask a variety of questions beyond only asking questions based on practices described in the document. 

Instead, crowdworkers are presented with public information about a mobile application available on the Google Play Store including its name, description and navigable screenshots. Figure \ref{fig:screenshots-question} shows an example of our user interface.\footnote{Color blindness affects approximately 4.5\% of the worlds population \cite{deeb2005molecular} We design all our crowdworker user interfaces to accommodate red-green color blindness.} Crowdworkers are asked to imagine they have access to a trusted third-party privacy assistant, to whom they can ask any privacy question about a given mobile application. We use the Amazon Mechanical Turk platform\footnote{\url{https://www.mturk.com/}} and recruit crowdworkers who have been conferred ``master'' status and are located within the United States of America. Turkers are asked to provide five questions per mobile application, and are paid \$2 per assignment, taking \textasciitilde eight minutes to complete the task. 


\subsection{Answer Selection}
To identify legally sound answers, we recruit seven experts with legal training to construct answers to Turker questions. Experts identify relevant evidence within the privacy policy, as well as provide meta-annotation on the question's relevance, subjectivity, OPP-115 category \citep{wilson2016creation}, and how likely any privacy policy is to contain the answer to the question asked. 


%
%


\begin{table}[] 
\resizebox{\columnwidth}{!}{
\begin{tabular}{lll} 
         \toprule
        Word  & (\%) & Example Question From \datasetname \\ \midrule 
\ Will & 13.03 & \makecell[l]{will my data be sold or available\\ to be used by third party entities?} \\ 
\ Do & 2.70 & do you sell any of my information \\ 
\ What & 1.84 & \makecell[l]{what is the worst case scenario of\\ giving my information to this game?} \\ 
\ Are & 1.60 & \makecell[l]{are there any safety concerns i\\ should know about using this app?} \\ 
\ \makecell[l]{Can/\\could} & 1.47 & can i control who sees my data? \\ 
\ How & 1.40 & how secure is my stored data? \\ 
\ Where & 1.14 & \makecell[l]{where is my account info and\\ online activity stored?} \\ 
\ Has & 0.69 &  has there ever been a security breach? \\ 
\ If & 0.61 & if i delete the app will you keep my data? \\ 
\ Does & 0.46 & \makecell[l]{does the app save the addresses\\ that i enter?} \\ 
\bottomrule
\end{tabular}
}
\caption{Ten most frequent first words in questions in the \textsc{PrivacyQA} dataset. We observe high lexical diversity in prefixes with 35 unique first word types and 131 unique combinations of first and second words. }
\label{table:firstword}
\end{table}

\subsection{Analysis}

 \begin{table*}[] 
 \resizebox{\linewidth}{!}{
\begin{tabular}{lll} 
 \toprule
Privacy Practice                     & Proportion & Example Question From \datasetname \\ \midrule 
First Party Collection/Use          &   41.9 \%         &  what data does this game collect?       \\ 
Third Party Sharing/Collection      &   24.5 \%          & will my data be sold to advertisers?       \\ 
Data Security                        &   10.5 \%          & how is my info protected from hackers?        \\ 
Data Retention                       &   4.1 \%        &  how long do you save my information?       \\ 
User Access, Edit and Deletion       &   2.0 \%       & can i delete my information permanently?        \\ 
User Choice/Control                 &   6.5 \%        & is there a way to opt out of data sharing        \\ 
Other                                &   4.8 \%        & does the app connect to the internet at any point?         \\ 
Policy Change                        &   0.2 \%        & where is the privacy statement        \\ 
International and Specific Audiences &   0.2 \%       & what are your GDPR policies?        \\ 
No Agreement                              &   5.4 \%        & how are features personalized?       \\ 
 \bottomrule
\end{tabular}
}
\caption{OPP-115 categories  most relevant to the questions collected from users.}
\label{opp:distribution}
 \end{table*}
\begin{table}[tb]
\begin{tabular}{llll}
        \toprule
Dataset              & Train & Test & All \\ \midrule
\# Questions          & 1350      & 400     & 1750    \\
 \# Policies           & 27      & 8     & 35    \\
 \# Sentences          & 3704      & 1243     & 4947   \\
 Avg Q. Length    & 8.42      & 8.56     & 8.46    \\
 Avg Doc. Length & 3121.3      & 3629.13     & 3237.37   \\
 Avg Ans. Length  & 123.73      & 153.44     & 139.62 \\
 \bottomrule
\end{tabular}
 \caption{Statistics of the \textsc{PrivacyQA} Dataset, where \# denotes number of questions, policies and sentences, and average length of questions, policies and answers in words, for training and test partitions.} 
 \label{data:statistics}
 \end{table}


 Table.\ref{data:statistics} presents aggregate statistics of the \textsc{PrivacyQA} dataset. 1750 questions are posed to our imaginary privacy assistant over 35 mobile applications and their associated privacy documents. As an initial step, we formulate the problem of answering user questions as an extractive sentence selection task, ignoring for now background knowledge, statistical data and legal expertise that could otherwise be brought to bear. 
 The dataset is partitioned into a training set featuring 27 mobile applications and 1350 questions, and a test set consisting of 400 questions over 8 policy documents. This ensures that documents in training and test splits are mutually exclusive. Every question is answered by at least one expert. In addition, in order to estimate annotation reliability and provide for better evaluation, every question in the test set is answered by at least two additional experts.

Table \ref{table:firstword} describes the distribution over first words of questions posed by crowdworkers. We also observe low redundancy in the questions posed by crowdworkers over each policy, with each policy receiving \textasciitilde 49.94 unique questions despite crowdworkers independently posing questions. Questions are on average 8.4 words long. As declining to answer a question can be a legally sound response but is seldom practically useful, answers to questions where a minority of experts abstain to answer are filtered from the dataset.  Privacy policies are \textasciitilde3000 words long on average. The answers to the question asked by the users typically have \textasciitilde100 words of evidence in the privacy policy document.

\subsubsection{Categories of Questions}
Questions are organized under nine categories from the OPP-115 Corpus annotation scheme \cite{wilson2016creation}:\\

\vspace{-7mm}
\begin{tight_enumerate}
\itemsep0em 
    \item First Party Collection/Use: What, why and how information is collected by the service provider
    \item Third Party Sharing/Collection: What, why and how information shared with or collected by third parties
    \item Data Security: Protection measures for user information
    \item Data Retention:  How long user information will be stored
    \item User Choice/Control: Control options available to users 
    \item User Access, Edit and Deletion: If/how users can access, edit or delete information
    \item Policy Change: Informing users if policy information has been changed
    \item International and Specific Audiences: Practices pertaining to a specific group of users
    \item Other: General text, contact information or practices not covered by other categories.
\end{tight_enumerate}

For each question, domain experts indicate one or more\footnote{For example, some questions such as \emph{'What information of mine is collected by this app and who is it shared with?'} can be identified as falling under both first party data/collection and third party collection/sharing categories.} relevant OPP-115 categories. We mark a category as relevant to a question if it is identified as such by at least two annotators. If no such category exists, the category is marked as `Other' if atleast one annotator has identified the `Other' category to be relevant. If neither of these conditions is satisfied, we label the question as having no agreement. The distribution of questions in the corpus across OPP-115 categories is as shown in Table.\ref{opp:distribution}. First party and third party related questions are the largest categories, forming nearly 66.4\% of all questions asked to the privacy assistant.



\subsubsection{Answer Validation}
When do experts disagree? We would like to analyze the reasons for potential disagreement on the annotation task, to ensure disagreements arise due to valid differences in opinion rather than lack of adequate specification in annotation guidelines. It is important to note that the annotators are experts rather than crowdworkers. Accordingly, their judgements can be considered valid, legally-informed opinions even when their perspectives differ.  For the sake of this question we randomly sample 100 instances in the test data and analyze them for likely reasons for disagreements. We consider a disagreement to have occurred when more than one expert does not agree with the majority consensus. By disagreement we mean there is no overlap between the text identified as relevant by one expert and another.

We find that the annotators agree on the answer for 74\% of the questions, even if the supporting evidence they identify is not identical i.e full overlap. They disagree on the remaining 26\%.  Sources of apparent disagreement correspond to situations when  different experts: have differing interpretations of question intent (11\%) (for example, when a user asks \emph{'who can contact me through the app'}, the questions admits multiple interpretations, including seeking information about the features of the app, asking about first party collection/use of data or asking about third party collection/use of data), identify different sources of evidence for questions that ask if a practice is performed or not (4\%), have differing interpretations of policy content (3\%), identify a partial answer to a question in the privacy policy (2\%) (for example, when the user asks \emph{`who is allowed to use the app'} a majority of our annotators decline to answer, but the remaining annotators highlight partial evidence in the privacy policy which states that children under the age of 13 are \emph{not} allowed to use the app), and other legitimate sources of disagreement (6\%) which include personal subjective views of the annotators (for example, when the user asks `is my DNA information used in any way \emph{other than what is specified}', some experts consider the boilerplate text of the privacy policy which states that it abides to practices described in the policy document as sufficient evidence to answer this question, whereas others do not).


\section{Experimental Setup}
We evaluate the ability of machine learning methods to identify relevant evidence for questions in the privacy domain.\footnote{The task of evidence identification can serve as a first step for future question answering systems, that can further learn to form abstractive summaries when required based on identifying relevant evidence.} We establish baselines for the subtask of deciding on the answerability (\S \ref{methods:answerability}) of a question, as well as the overall task of identifying evidence for questions from policies (\S \ref{methods:baselines}). We describe aspects of the question that can render it unanswerable within the privacy domain (\S \ref{ref:unanswerablefactors}).

\begin{table}[]
\centering
\resizebox{\columnwidth}{!}{
\begin{tabular}{|l|l|l|l|l|}
\hline
    & Acc. & P & R & F1 \\ \hline
Majority  &   24.75      & 24.75  &  100 &  39.6  \\ \hline
SVM-BOW  &    75.75      & 50.8  &  58.5  &  54.4 \\ 
+ LEN  &    76.75      & 52.7  &  57.5  &  55.0 \\ 
+ LEN + POS &    77.0      & 53.2  &  58.5  &  55.7 \\ \hline
CNN &       80.0   & 61.1  & 52.5  & 56.5   \\ \hline
BERT & 81.15 & 62.6 & 62.6 & 62.6 \\ \hline
\end{tabular}
}
\caption{Classifier Performance (\%) for answerability of questions. The Majority Class baseline always predicts that questions are unanswerable.}
\label{table:unanswerable}
\end{table}

\begin{table}[t]
\resizebox{\columnwidth}{!}{
\begin{tabular}{|l|l|l|l|l|}
\hline
Model      & Precision & Recall & F1    \\ \hline
No Answer (NA)  & 28.0\%    & 28.0\% & 28.0\%  \\ \hline
Word Count -2 & 24.0\%    & 16.4\% & 19.4\%  \\ 
Word Count -3 & 21.8\%    & 17.8\% & 19.6\%  \\ 
Word Count -5 & 18.1\%    & 20.3\% & 19.2\%  \\  \hline

BERT & 44.2\% & 34.8\%  & 39.0\%   \\ 
BERT + Unans. & 44.3\% & 36.1\%  & \textbf{39.8}\%    \\ \hline
Human       & 68.8\% & 69.0\% & 68.9\%       \\ \hline
\end{tabular}
}
\caption{Performance of baselines on \textsc{PrivacyQA} dataset.}
\label{table:baselines}
\end{table}
\subsection{Answerability Identification Baselines}
\label{methods:answerability}

\begin{table}[t] 
\resizebox{\columnwidth}{!}{
\begin{tabular}{|llll|} 
 \toprule
Privacy Practice                 & NA  & BERT-U     & Human  \\ \midrule 
First Party        &   0.22   &   0.36    &   0.67     \\ 
Third Party  &   0.10   &   0.26    &   0.61     \\ 
Data Security                    &   0.24    &   0.42  &   0.74        \\ 
Data Retention                   &   0.02    &   0.33  &   0.67   \\ 
User Access  &   0.07    &   0.32  &   0.66   \\ 
User Choice              &   0.35   &   0.41  &   0.65   \\ 
Other                            &   0.44    &   0.45  &   0.72    \\ 
 \bottomrule
\end{tabular}
}
\caption{Stratification of classifier performance by OPP-115 category of questions.}
\label{table:oppstrata}
 \end{table}
 
 \begin{table}[]
\centering
\begin{tabular}{|l|l|}
\hline
                                         & BERT  \\ \hline
\# Answerability Mistakes                &   137               \\ \hline
\% Answerable -\textgreater Unanswerable & 124                  \\ \hline
\% Unanswerable -\textgreater Answerable &  13               \\ \hline
Out-of-scope                             &  2                \\ \hline
Subjective                               &    46              \\ \hline
Policy Silent                            &    19              \\ \hline
Unexpected                               &     6             \\ \hline
\end{tabular}
\caption{Analysis of BERT performance at identifying answerability. The majority of mistakes made by BERT are answerable Questions identified as unanswerable. These answerable questions are further analyzed along the factors of scope, subjectivity, presence of answer and whether the question could be anticipated.}
\end{table}

We define answerability identification as a binary classification task, evaluating model ability to predict if a question can be answered, given a question in isolation. This can serve as a prior for downstream question-answering.   We describe three baselines on the answerability task, and find they considerably improve performance over a majority-class baseline.

\textbf{SVM}: We define 3 sets of features to characterize each question. The first is a simple bag-of-words set of features over the question (\textsc{SVM-BOW}), the second is bag-of-words features of the question as well as length of the question in words (\textsc{SVM-BOW + LEN}), and lastly we extract bag-of-words features, length of the question in words as well as part-of-speech tags for the question (\textsc{SVM-BOW + LEN + POS}). This results in vectors of 200, 201 and 228 dimensions respectively, which are provided to an SVM with a linear kernel.

\textbf{CNN}: We utilize a CNN neural encoder for answerability prediction. We use GloVe word embeddings \cite{pennington2014glove}, and a filter size of 5 with 64 filters to encode questions.

\textbf{BERT}: BERT \cite{devlin2018bert} is a bidirectional transformer-based language-model \cite{vaswani2017attention}.\footnote{We utilize the HuggingFace implementtation available at \url{https://github.com/huggingface/pytorch-transformers}} We fine-tune BERT-base on our binary answerability identification task with a learning rate of 2e-5 for 3 epochs, with a maximum sequence length of 128.
\label{section:answerability}

\subsection{Privacy Question Answering}
\label{methods:baselines}
Our goal is to identify evidence within a privacy policy for questions asked by a user. This is framed as an answer sentence selection task, where models identify a set of evidence sentences from all candidate sentences in each policy. 
\subsubsection{Evaluation Metric}
Our evaluation metric for answer-sentence selection is sentence-level F1, implemented similar to \cite{choi2018quac, rajpurkar2016squad}. Precision and recall are implemented by measuring the overlap between predicted sentences and sets of gold-reference sentences. We report the average of the maximum F1 from each n$-$1 subset, in relation to the heldout reference.

\subsubsection{Baselines}
We describe baselines on this task, including a human performance baseline.

\textbf{No-Answer Baseline (NA)} : Most of the questions we receive are difficult to answer in a legally-sound way on the basis of information present in the privacy policy. We establish a simple baseline to quantify the effect of identifying every question as unanswerable.




\textbf{Word Count Baseline} : To quantify the effect of using simple lexical matching to answer the questions, we retrieve the top candidate policy sentences for each question using a  word count baseline \cite{yang2015wikiqa}, which counts the number of question words that also appear in a sentence. We include the top 2, 3 and 5 candidates as baselines.  

\textbf{BERT}: We implement two BERT-based baselines \cite{devlin2018bert} for evidence identification. First, we train BERT on each query-policy sentence pair as a binary classification task to identify if the sentence is evidence for the question or not (\textsc{Bert}). We also experiment with a two-stage classifier, where we separately train the model on questions only to predict answerability. At inference time, if the answerable classifier predicts the question is answerable, the evidence identification classifier produces a set of candidate sentences (\textsc{Bert + Unanswerable}).

\textbf{Human Performance}: We pick each reference answer provided by an annotator, and compute the F1 with respect to the remaining references, as described in section 4.2.1. Each reference answer is treated as the prediction, and the remaining n-1 answers are treated as the gold reference. The average of the maximum F1 across all reference answers is computed as the human baseline.
\section{Results and Discussion}
The results of the answerability baselines are presented in Table \ref{table:unanswerable}, and on answer sentence selection in  Table \ref{table:baselines}. We observe that \textsc{bert} exhibits the best performance on a binary answerability identification task. However, most baselines considerably exceed the performance of a majority-class baseline. This suggests considerable information in the question, indicating it's possible answerability within this domain.

Table.\ref{table:baselines} describes the performance of our baselines on the answer sentence selection task. The No-answer (NA) baseline performs at 28 F1, providing a lower bound on performance at this task. We observe that our best-performing baseline, \textsc{Bert + Unanswerable} achieves an F1 of 39.8. This suggest that \textsc{bert} is capable of making some progress towards answering questions in this difficult domain, while still leaving considerable headroom for improvement to reach human performance. \textsc{Bert + Unanswerable} performance suggests that incorporating information about answerability can help in this difficult domain. We examine this challenging phenomena of unanswerability further in Section \ref{unanswerablefactors}.
\subsection{Error Analysis}
\label{analysis}

Disagreements are analyzed based on the OPP-115 categories of each question (Table.\ref{table:oppstrata}). We compare our best performing BERT variant against the NA model and human performance. We observe significant room for improvement across all categories of questions but especially for first party, third party and data retention categories.


We analyze the performance of our strongest BERT variant, to identify classes of errors and directions for future improvement (Table.8). We observe that a majority of answerability mistakes made by the BERT model are questions which are in fact answerable, but are identified as unanswerable by BERT. We observe that BERT makes 124 such mistakes on the test set. We collect expert judgments on relevance, subjectivity , silence and information about how likely the question is to be answered from the privacy policy from our experts. We find that most of these mistakes are relevant questions. However many of them were identified as subjective by the annotators, and at least one annotator marked 19 of these questions as having no answer within the privacy policy. However, only 6 of these questions were unexpected or do not usually have an answer in privacy policies. These findings suggest that a more nuanced understanding of answerability might help improve model performance in his challenging domain.
\subsection{What makes Questions Unanswerable?}
\label{ref:unanswerablefactors}

We further ask legal experts to identify potential causes of unanswerability of questions. This analysis has considerable implications. While past work \cite{rajpurkar2018know} has treated unanswerable questions as homogeneous,  a question answering system might wish to have different treatments for different categories of `unanswerable' questions. The following factors were identified to play a role in unanswerability:

\begin{itemize}
\itemsep0em 
    \item \textbf{Incomprehensibility}: If a question is incomprehensible to the extent that its meaning is not intelligible. 
    \item \textbf{Relevance}: Is this question in the scope of what could be answered by reading the privacy policy.
    \item \textbf{Ill-formedness}: Is this question ambiguous or vague. An ambiguous statement will typically contain expressions that can refer to multiple potential explanations, whereas a vague statement carries a concept with an unclear or soft definition.
\item \textbf{Silence}: Other policies answer this type of question but this one does not.
\item  \textbf{Atypicality}: The question is of a nature such that it is unlikely for any policy policy to have an answer to the question.
\end{itemize}
Our experts attempt to identify the different `unanswerable' factors for all 573 such questions in the corpus. 4.18\% of the questions were identified as being incomprehensible (for example, `any difficulties to occupy the privacy assistant'). Amongst the comprehendable questions, 50\% were identified as likely to have an answer within the privacy policy, 33.1\% were identified as being privacy-related questions but not within the scope of a privacy policy (e.g., \emph{'has Viber had any privacy breaches in the past?'}) and 16.9\% of questions were identified as completely out-of-scope (e.g., \emph{`'will the app consume much space?'}). In the questions identified as relevant, 32\% were ill-formed questions that were phrased by the user in a manner considered vague or ambiguous. Of the questions that were both relevant as well as `well-formed', 95.7\% of the questions were not answered by the policy in question but it was reasonable to expect that a privacy policy would contain an answer. The remaining 4.3\% were described as reasonable questions, but of a nature generally not discussed in privacy policies. This suggests that the answerability of questions over privacy policies is a complex issue, and future systems should consider each of these factors when serving user's information seeking intent.




We examine a large-scale dataset of ``natural'' unanswerable questions \cite{kwiatkowski2019natural} based on real user search engine queries to identify if similar unanswerability factors exist. It is important to note that these questions have previously been filtered, according to a criteria for bad questions defined as  ``(questions that are)  ambiguous,  incomprehensible,  dependent  on clear false presuppositions, opinion-seeking, or not clearly a request for factual information.'' Annotators made the decision based on the content of the question without viewing the equivalent Wikipedia page. We randomly sample 100 questions from the development set which were identified as unanswerable, and find that 20\% of the questions are not questions (e.g., ``all I want for christmas is you mariah carey tour''). 12\% of questions are unlikely to ever contain an answer on Wikipedia, corresponding closely to our atypicality category. 3\% of questions are unlikely to have an answer anywhere (e.g., `what guides Santa home after he has delivered presents?'). 7\% of questions are incomplete or open-ended (e.g., `the south west wind blows across nigeria between'). 3\% of questions have an unresolvable coreference  (e.g., `how do i get to Warsaw Missouri from here'). 4\% of questions are vague, and a further 7\% have unknown sources of error. 2\% still contain false presuppositions (e.g., `what is the only fruit that does not have seeds?') and the remaining 42\% do not have an answer within the document.  This reinforces our belief that though they have been understudied in past work, any question answering system interacting with real users should expect to receive such unanticipated and unanswerable questions.
\section{Conclusion}
We present \datasetname, the first significant corpus of privacy policy questions and more than 3500 expert annotations of relevant answers. The goal of this work is to promote question-answering research in the specialized privacy domain, where it can have large real-world impact. Strong neural baselines on \textsc{PrivacyQA} achieve a performance of only 39.8 F1 on this corpus, indicating considerable room for future research. Further, we shed light on several important considerations that affect the \emph{answerability} of questions. We hope this contribution leads to multidisciplinary efforts  to precisely understand user intent and reconcile it with information in policy documents, from both the privacy and NLP communities. 
\section*{Acknowledgements}
This research was supported in part by grants from the National Science Foundation Secure and Trustworthy Computing program (CNS-1330596, CNS-1330214, CNS-15-13957, CNS-1801316, CNS-1914486, CNS-1914444) and a DARPA Brandeis grant on Personalized Privacy Assistants (FA8750-15-2-0277). The US Government is authorized to reproduce and distribute reprints for Governmental purposes not withstanding any copyright notation. The views and conclusions contained herein are those of the authors and should not be interpreted as necessarily representing the official policies or endorsements, either expressed or implied, of the NSF, DARPA, or the US Government. The authors would like to extend their gratitude to Elias Wright, Gian Mascioli, Kiara Pillay, Harrison Kay, Eliel Talo, Alexander Fagella and N. Cameron Russell for providing their valuable expertise and insight to this effort. The authors are also grateful to Eduard Hovy, Lorrie Cranor, Florian Schaub, Joel Reidenberg, Aditya Potukuchi and Igor Shalyminov for helpful discussions related to this work, and to the three anonymous reviewers of this draft for their constructive feedback. Finally, the authors would like to thank all crowdworkers who consented to participate in this study.

\bibliography{emnlp-ijcnlp-2019}
\bibliographystyle{acl_natbib}

\appendix

\end{document}